\documentclass{article} 
\usepackage{iclr2017_conference,times}
\usepackage{array}
\usepackage{graphicx}
\usepackage{url}
\usepackage{multicol}
\usepackage[ruled,linesnumbered]{algorithm2e}
\usepackage{subcaption}
\usepackage{amsmath}
\usepackage{wrapfig}

\newcommand*\samethanks[1][\value{footnote}]{\footnotemark[#1]}

\title{Revisiting Distributed Synchronous SGD}

\author{Jianmin Chen\thanks{Joint first authors}, Xinghao Pan\samethanks[1]~\thanks{UC Berkeley, Berkeley, CA, USA, \texttt{xinghao@eecs.berkeley.edu}}, Rajat Monga, Samy Bengio\\
Google Brain\\
Mountain View, CA, USA\\
\texttt{\{jmchen,xinghao,rajatmonga,bengio\}@google.com}\\
\And
Rafal Jozefowicz \\
OpenAI\\
San Francisco, CA, USA\\
\texttt{rafal@openai.com}
}

\begin{document}

\maketitle

%

\begin{abstract}
Distributed training of deep learning models on large-scale training data is typically conducted with {\em asynchronous} stochastic optimization to maximize the rate of updates, at the cost of additional noise introduced from asynchrony.
In contrast, the {\em synchronous} approach is often thought to be impractical due to idle time wasted on waiting for straggling workers.
We revisit these conventional beliefs in this paper, and examine the weaknesses of both approaches.
We demonstrate that a third approach, synchronous optimization with backup workers, can avoid asynchronous noise while mitigating for the worst stragglers.
Our approach is empirically validated
and shown to converge {\em faster} and to {\em better} test accuracies.
\end{abstract}

\section{Introduction}
\label{sec:intro}
The recent success of deep learning approaches for domains like speech recognition~\citep{dnn-speech} and computer vision~\citep{batchnorm} stems from many algorithmic improvements but also from the fact that the size of available training data has grown significantly over the years, together with the computing power, in terms of both CPUs and GPUs.
While a single GPU often provides algorithmic simplicity and speed up to a given scale of data and model, there exist an operating point where a distributed implementation of training algorithms for deep architectures becomes necessary.

Currently, popular distributed training algorithms include mini-batch versions of stochastic gradient descent (SGD) and other stochastic optimization algorithms such as AdaGrad \citep{duchi2011adaptive}, RMSProp \citep{tieleman2012lecture}, and ADAM \citep{kingma2014adam}.
Unfortunately, bulk-synchronous implementations of stochastic optimization are often slow in practice due to the need to wait for the slowest machine in each synchronous batch.
To circumvent this problem, practitioners have resorted to asynchronous approaches which emphasize speed by using potentially stale information for computation.
While asynchronous training have proven to be faster than their synchronous counterparts, they often result in convergence to poorer results.

In this paper\footnote{This is an extension of our ICLR 2016 workshop extended abstract \citep{chen2016revisiting}.}, we revisit synchronous learning, and propose a method for mitigating stragglers in synchronous stochastic optimization.
Specifically, we synchronously compute a mini-batch gradient with only a subset of worker machines, thus alleviating the straggler effect while avoiding any staleness in our gradients.
The primary contributions of our paper are:
\begin{itemize}
\item Illustration of how gradient staleness in asynchronous training negatively impacts test accuracy and is exacerbated by deep models.
\item Measurement of machine response times for synchronous stochastic optimization in a large deployment of 100 GPUs, showing how stragglers in the tail end affect convergence speed.
\item Proposal of synchronous stochastic optimization with {\em backup workers} to mitigate straggler effects without gradient staleness.
\item Establishing the need to measure both speed of convergence and test accuracy of optimum for empirical validation.
\item Empirical demonstration that our proposed synchronous training method outperforms asynchronous training by converging faster and to better test accuracies.
\end{itemize}


The remainder of this paper is organized as follows.
We briefly present preliminaries and notation in Section \ref{sec:prelims}.
Section \ref{sec:asyncsgd} describes asynchronous stochastic optimization and presents experimental evidence of gradient staleness in deep neural network models.
We present our approach in Section \ref{sec:syncsgd}, and exhibit straggler effects that motivate the approach.
We then empirically evaluate our approach in Sections \ref{sec:expt}.
Related work is discussed in Section \ref{sec:related}, and we conclude in Section \ref{sec:conclude}.

\subsection{Preliminaries and Notation}
\label{sec:prelims}
Given a dataset $\mathcal{X} = \{x_i : i = 1,\dots, |\mathcal{X}|\}$, our goal is to learn the parameters $\theta$ of a model with respect to an empirical loss function $f$, defined as
$f(\theta) \overset{\Delta}{=} \frac{1}{|\mathcal{X}|} \sum_{i=1}^{|\mathcal{X}|} F(x_i; \theta)$,
where $F(x_i;\theta)$ is the loss with respect to a datapoint $x_i$ and the model $\theta$.

A first-order stochastic optimization algorithm achieves this by iteratively updating $\theta$ using a stochastic gradient $G \overset{\Delta}{=} \nabla F(x_i; \theta)$ computed at a randomly sampled $x_i$, producing a sequence of models $\theta^{(0)}, \theta^{(1)}, \dots$.
Stochastic optimization algorithms differ in their update equations.
For example, the update of SGD is $\theta^{(t+1)} = \theta^{(t)} - \gamma_t G^{(t)} = \theta^{(t)} - \gamma_t \nabla F(x_i; \theta^{(t)})$, where $\gamma_t$ is the {\em learning rate} or {\em step size} at iteration $t$.
A mini-batch version of the stochastic optimization algorithm computes the stochastic gradient over mini-batch of size $B$ instead of a single datapoint, i.e., $G \overset{\Delta}{=} \frac{1}{B} \sum_{i=1}^B \nabla F(\widetilde{x}_i; \theta^{(t)})$, where $\widetilde{x}_i$'s are randomly sampled from $\mathcal{X}$.
We will often evaluate performance on an exponential moving average $\bar\theta^{(t)} = \alpha \bar\theta^{(t-1)} + (1 - \alpha) \theta^{(t)}$ with decay rate $\alpha$.

Our interest is in {\em distributed} stochastic optimization using $N$ worker machines in charge of computing stochastic gradients that are sent to $M$ parameter servers.
Each parameter server $j$ is responsible for storing a subset $\theta[j]$ of the model, and performing updates on $\theta[j]$.
In the synchronous setting, we will also introduce additional $b$ {\em backup workers} for straggler mitigation.

\section{Asynchronous Stochastic Optimization}
\label{sec:asyncsgd}

An approach for a distributed
stochastic gradient descent algorithm was presented in \cite{dean:2012}, consisting of two main
ingredients. First, the parameters of the model are distributed
on multiple servers, depending on the architecture. This set of servers are called
the {\em parameter servers}. Second, there can be multiple workers
processing data in parallel and communicating with the parameter servers.
Each worker processes a mini-batch of data independently of the others,
as follows:
\begin{itemize}
\item The worker fetches from the parameter servers the most up-to-date
parameters of the model needed to process the current mini-batch;
\item It then computes gradients of the loss with respect to these
parameters;
\item Finally, these gradients are sent back to the parameter servers, which
then updates the model accordingly.
\end{itemize}
Since each worker communicates with the parameter servers independently of the others,
this is called {\em Asynchronous Stochastic Gradient Descent} (Async-SGD), or more generally, {\em Asynchronous Stochastic Optimization} (Async-Opt).
A similar approach was later proposed by~\cite{adam:2014}.
Async-Opt is presented in Algorithms \ref{alg:async_worker} and \ref{alg:async_ps}.

\begin{figure*}[ht]
\begin{multicols}{2}
\small
\begin{algorithm}[H]
\caption{Async-SGD worker $k$}
\label{alg:async_worker}
\DontPrintSemicolon
\SetKwInOut{Input}{Input}
\Input{Dataset $\mathcal{X}$}
\Input{$B$ mini-batch size}
\While{True}{
  Read $\widehat\theta_k = (\theta[0], \dots, \theta[M])$ from PSs.\label{alg:async_worker:read}\;
  $G_k^{(t)} := 0$.\;
  \For{$i = 1,\dots,B$}{
    Sample datapoint $\widetilde{x}_i$ from $\mathcal{X}$.\;
    $G_k^{(t)} \leftarrow G_k^{(t)} + \frac{1}{B} \nabla F(\widetilde{x}_i; \widehat\theta_k)$.
  }
  Send $G_k^{(t)}$ to parameter servers.
}
\end{algorithm}

\columnbreak

\begin{algorithm}[H]
\caption{Async-SGD Parameter Server $j$}
\label{alg:async_ps}
\DontPrintSemicolon
\SetKwInOut{Input}{Input}
\Input{$\gamma_0,\gamma_1,\dots$ learning rates.}
\Input{$\alpha$ decay rate.}
\Input{$\theta^{(0)}$ model initialization.}
\For{$t = 0, 1, \dots$}{
  Wait for gradient $G$ from any worker.\;
  $\theta^{(t+1)}[j] \leftarrow \theta^{(t)}[j] - \gamma_t G[j]$.\;
  $\bar\theta^{(t)}[j] = \alpha \bar\theta^{(t-1)}[j] + (1 - \alpha) \theta^{(t)}[j]$.
}
\end{algorithm}
\end{multicols}
\end{figure*}

In practice, the updates of Async-Opt are different than those of serially running the stochastic optimization algorithm for two reasons.
Firstly, the read operation (Algo \ref{alg:async_worker} Line \ref{alg:async_worker:read}) on a worker may be interleaved with updates by other workers to different parameter servers, so the resultant $\widehat\theta_k$ may not be consistent with any parameter incarnation $\theta^{(t)}$.
Secondly, model updates may have occurred while a worker is computing its stochastic gradient;
hence, the resultant gradients are typically computed with respect to outdated parameters.
We refer to these as {\em stale} gradients, and its {\em staleness} as the number of updates that have occurred between its corresponding read and update operations.

Understanding the theoretical impact of staleness is difficult work and the topic of many recent papers, e.g. \cite{recht2011hogwild, duchi2013estimation, leblond2016asaga, reddi2015variance, de2015taming, mania2015perturbed}, most of which focus on individual algorithms, under strong assumptions that may not hold up in practice.
This is further complicated by deep models with multiple layers, since the times at which model parameters are read and which gradients are computed and sent are dependent on the depth of the layers (Figure \ref{fig:gradient_flow_layers}).
To better understand this dependence in real models, we collected staleness statistics on a Async-Opt run with 40 workers on a 18-layer Inception model \citep{inception_v3} trained on the ImageNet Challenge dataset \citep{imagenet}, as shown in Table \ref{tab:inceptionstalenss}.


\begin{wraptable}{r}{8.5cm}
\small
\vspace{-0.75cm}
\begin{tabular}{|c||c|c|c|c||c|c|}
\hline
Layer & Min & Mean & Median & Max & Std Dev & Count\\\hline
        18 &          4 &      14.54 &      13.94 &         29 &       3.83 &      10908 \\
        12 &          5 &      11.35 &       11.3 &         23 &       3.09 &      44478 \\
        11 &          8 &       19.8 &      19.59 &         34 &       3.65 &        187 \\
         0 &         24 &      38.97 &      38.43 &         61 &       5.43 &        178 \\
\hline
\end{tabular}
\caption{\small Staleness of gradients in a 18-layer Inception model.
Gradients were collected in a run of asynchronous training using 40 machines.
{\em Staleness} of a gradient is measured as the number of updates that have occurred between its corresponding read and update operations.
The staleness of gradients increases from a mean of $\sim$14.5 in the top layer (Layer 18) to $\sim$39.0 in the bottom layer (Layer 0).}
\label{tab:inceptionstalenss}
\vspace{-1.5cm}
\end{wraptable}

Despite the abovementioned problems, Async-Opt has been shown to be scale well up to a few dozens of workers for some models.
However, at larger scales, increasing the number of machines (and thus staleness of gradients) can result in poorer trained models.

\subsection{Impact of staleness on test accuracy}
\label{sec:asyncsgd:staleness}
We explore how increased staleness contributes to training of poorer models.
In order to mimic the setting on a smaller scale, we trained a
state-of-the-art MNIST CNN model but simulated staleness by using old gradients for the parameter
updates.
Details of the model and training are provided in Appendix \ref{app:model:mnistcnn}.

\begin{figure}
\centering
\begin{minipage}{0.64\textwidth}
\centering
\includegraphics[width=0.99\textwidth]{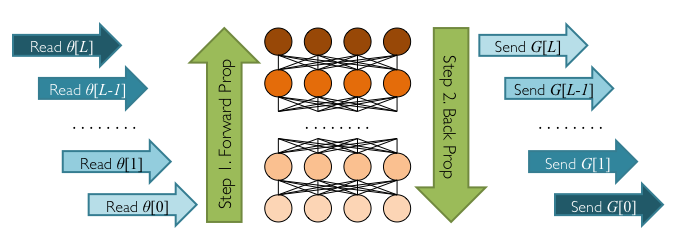}
\caption{\small Gradient staleness dependence on model layer.
Gradients are computed in a bottom-up forward propagation step followed by a top-down back propagation step.
Parameters are read from servers in the forward prop, but gradients are sent to servers during the back prop.
Thus, gradients of lower layers are more stale than top layers.}
\label{fig:gradient_flow_layers}
\end{minipage}
\hspace{0.1cm}
\begin{minipage}{0.34\textwidth}
\centering
\includegraphics[width=0.99\textwidth]{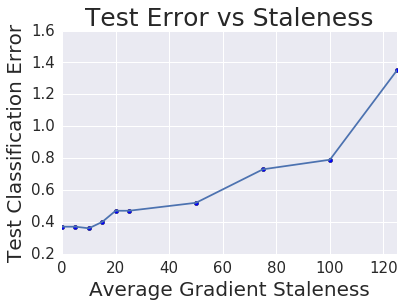}
\caption{\small Degradation of test classification error with increasing average gradient staleness in MNIST CNN model.}
\label{fig:gradstalemnist}
\end{minipage}
\end{figure}

The best final classification error on a test set was 0.36\%, which increases to
0.47\% with average gradient staleness of 20 steps, and up to 0.79\% with 50
steps (see Figure \ref{fig:gradstalemnist}).

Once the average simulated staleness was chosen to be more than 15 steps, the results
started to significantly deteriorate and the training itself became much less
stable. We had to employ following tricks to prevent the results from blowing up:
\begin{itemize}
\item Slowly increase the staleness over the first 3 epochs of training. This
  mimics increasing the number of asynchronous workers and is also very important
    in practice for some of the models we experimented with (e.g. large
    word-level language models). The trick was not relevant with a simulated staleness less than
    15 but became crucial for larger values.
\item Use lower initial learning rates when staleness is at least 20, which reduces
  a frequency of explosions (train error goes to 90\%). This observation is
    similar to what we found in other experiments - we were able to use much
    larger learning rates with synchronous training and the results were also
    more stable.
\item Even with above tricks the divergence occurs occasionally and we found
  that restarting training from random weights can lead to more successful runs.
    The best results were then chosen based on validation set performance.
\end{itemize}

\section{Revisting Synchronous Stochastic Optimization}
\label{sec:syncsgd}
Both~\cite{dean:2012} and~\cite{adam:2014} use versions of Async-SGD where
the main potential problem is that each worker computes gradients
over a potentially old version of the model.
In order to remove this discrepancy, we propose here to reconsider
a synchronous version of distributed stochastic gradient descent (Sync-SGD),
or more generally, {\em Synchronous Stochastic Optimization} (Sync-Opt),
where the parameter servers wait for all workers to send their
gradients, aggregate them, and send the updated parameters to all workers
afterward.
This ensures that the actual algorithm is a true mini-batch
stochastic gradient descent, with an effective batch size equal to the sum of
all the mini-batch sizes of the workers.

While this approach solves the staleness problem, it also introduces
the potential problem that the actual update time now depends on the slowest worker.
Although workers have equivalent computation and network communication workload,
slow stragglers may result from failing hardware, or contention on shared underlying hardware resources in data centers, or even due to preemption by other jobs.

To alleviate the straggler problem, we introduce {\em backup workers} ~\citep{tail-at-scale}
as follows: instead of having only $N$ workers, we add $b$ extra workers,
but as soon as the parameter servers receive gradients from any $N$
workers, they stop waiting and update their parameters using the $N$ gradients.
The slowest $b$ workers' gradients will be dropped when they arrive.
Our method is presented in Algorithms \ref{alg:sync_worker}, \ref{alg:sync_ps}.

\begin{figure*}[h]
\begin{multicols}{2}
\small
\begin{algorithm}[H]
\caption{Sync-SGD worker $k$, where $k=1,\dots,N+b$}
\label{alg:sync_worker}
\DontPrintSemicolon
\SetKwInOut{Input}{Input}
\Input{Dataset $\mathcal{X}$}
\Input{$B$ mini-batch size}
\For{$t = 0,1,\dots$}{
  Wait to read $\theta^{(t)} = (\theta^{(t)}[0], \dots, \theta^{(t)}[M])$ $\quad\quad$from parameter servers.\;
  $G_k^{(t)} := 0$\;
  \For{$i = 1,\dots,B$}{
    Sample datapoint $\widetilde{x}_{k,i}$ from $\mathcal{X}$.\;
    $G_k^{(t)} \leftarrow G_k^{(t)} + \frac{1}{B} \nabla F(\widetilde{x}_{k,i}; \theta^{(t)})$.
  }
  Send $(G_k^{(t)}, t)$ to parameter servers.
}
\end{algorithm}

\columnbreak

\begin{algorithm}[H]
\caption{Sync-SGD Parameter Server $j$}
\label{alg:sync_ps}
\DontPrintSemicolon
\SetKwInOut{Input}{Input}
\Input{$\gamma_0,\gamma_1,\dots$ learning rates.}
\Input{$\alpha$ decay rate.}
\Input{$N$ number of mini-batches to aggregate.}
\Input{$\theta^{(0)}$ model initialization.}
\For{$t = 0, 1, \dots$}{
  $\mathcal{G} = \{\}$\;
  \While{$|\mathcal{G}| < N$}{
    Wait for $(G, t')$ from any worker.\;
    \lIf{$t' == t$}{$\mathcal{G} \leftarrow \mathcal{G} \cup \{G\}$.}
    \lElse{Drop gradient $G$.}
  }
  $\theta^{(t+1)}[j] \leftarrow \theta^{(t)}[j] - \frac{\gamma_t}{N} \sum_{G\in\mathcal{G}} G[j]$.\;
  $\bar\theta^{(t)}[j] = \alpha \bar\theta^{(t-1)}[j] + (1 - \alpha) \theta^{(t)}[j]$.
}
\end{algorithm}
\end{multicols}
\end{figure*}


\subsection{Straggler effects}
\label{sec:syncsgd:straggler}
The use of backup workers is motivated by the need to mitigate slow stragglers while maximizing computation.
We investigate the effect of stragglers on Sync-Opt model training here.

\begin{figure}[h]
\centering
\begin{minipage}[t]{0.49\textwidth}
\centering
\includegraphics[width=0.99\textwidth]{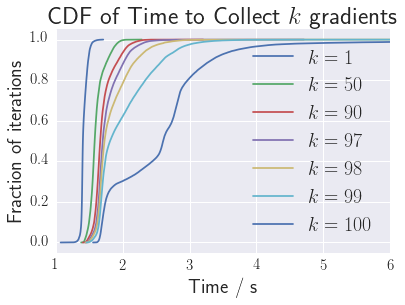}
\caption{\footnotesize CDF of time taken to aggregate gradients from $N$ machines.
For clarity, we only show times of $\leq 6$s; the maximum observed time is 310s.}
\label{fig:straggler_cdf}
\end{minipage}
\hspace{0.1cm}
\begin{minipage}[t]{0.49\textwidth}
\centering
\includegraphics[width=0.99\textwidth]{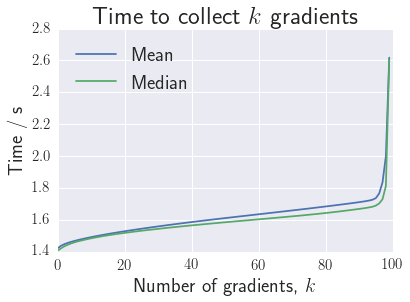}
\caption{\footnotesize Mean and median times, across all iterations, to collect $k$ gradients on $N=100$ workers and $b=0$ backups.
Most mean times fall between 1.4s and 1.8s, except of final few gradients.}
\label{fig:straggler_mean}
\end{minipage}
\end{figure}

We ran Sync-Opt with $N=100$ workers, $b=0$ backups, and 19 parameter servers on the Inception model.
Using one variable as a proxy, we collected for each iteration both the start time of the iteration and the time when the $k$th gradient of that variable arrived at the parameter server.
These times are presented in Figure \ref{fig:straggler_cdf} for $k=1,50,90,97,98, 99, 100$.
Note that 80\% of the 98th gradient arrives in under 2s, whereas only 30\% of the final gradient do.
Furthermore, the time to collect the final few gradients grows exponentially, resulting in wasted idle resources and time expended to wait for the slowest gradients.
This exponential increase is also seen in Figure \ref{fig:straggler_mean}.

Thus, one might choose to drop slow stragglers to decrease the iteration time.
However, using fewer machines implies a smaller effective mini-batch size and thus greater gradient variance, which in turn could require more iterations for convergence.
We examine this relationship by running Sync-Opt\footnote{
Since we are interested in the gradient quality and convergence behavior but {\em not} running time in this experiment, the backups serve only to reduce our data collection time but do not affect our analysis.
}
with $N=50, 70, 80, 90, 100$ and $b=6$, and note the number of iterations required for convergence in Figure \ref{fig:straggler_iterations}.
Additional details of this training are provided in Appendix \ref{app:model:straggler_inception}.
As $N$ is doubled from 50 to 100, the number of iterations to converge nearly halves from $137.5e3$ to $76.2e3$.

\begin{figure}[h]
\centering
\begin{minipage}[t]{0.49\textwidth}
\centering
\includegraphics[width=0.99\textwidth]{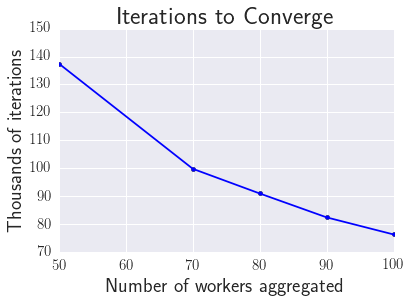}
\caption{\footnotesize Number of iterations to converge when aggregating gradient from $N$ machines.}
\label{fig:straggler_iterations}
\end{minipage}
\hspace{0.1cm}
\begin{minipage}[t]{0.49\textwidth}
\centering
\includegraphics[width=0.99\textwidth]{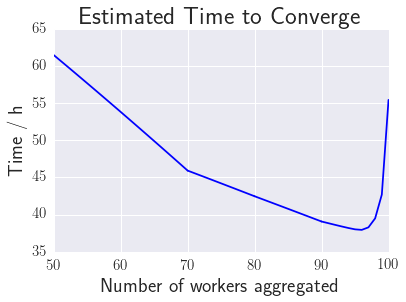}
\caption{\footnotesize Estimated time to converge when aggregating gradients from $N$ machines on a $N+b=100$ machine configuration.
Convergence is fastest when choosing $N=96$, $b=4$.
}
\label{fig:straggler_esttime}
\end{minipage}
\end{figure}


Hence, there is a trade-off between dropping more stragglers to reduce iteration time, and waiting for more gradients to improve the gradient quality.
Consider a hypothetical setting where we have $N+b=100$ machines, and we wish to choose the best configuration of $N$ and $b$ to minimize running time to convergence.
For each configuration, we can estimate the iterations required from Figure \ref{fig:straggler_iterations} (linearly interpolating for values of $N$ for which we did not collect data).
We can multiply this with the mean iteration times (Figure \ref{fig:straggler_mean}) to obtain the {\em running time} required to converge for each setting of $N$ and $b$.
These results are shown in Figure \ref{fig:straggler_esttime}, indicating that $N=96$, $b=4$ converges fastest.
Therefore, this motivates our choice to use a few backup workers for mitigating stragglers.

\section{Experiments}
\label{sec:expt}

In this section, we present our empirical comparisons of synchronous and asynchronous distributed stochastic optimization algorithms
as applied to models such as Inception and PixelCNN.
All experiments in this paper are using the TensorFlow system \citep{tensorflow2015-whitepaper}.



\subsection{Metrics of comparison: Faster convergence, Better optimum}
\label{sec:expt:metrics}
We are interested in two metrics of comparison for our empirical validation:
(1) test error or accuracy, and (2) speed of convergence.
We point out that for non-convex deep learning models, it is possible to converge faster to a poorer local optimum.
Here we show a simple example with Inception using different learning rates.

\begin{table}[ht]
\begin{minipage}[b]{0.34\textwidth}
\small
\begin{tabular}[t]{|c||c|c|}\hline
Initial & Test & Epochs\\
rate & precision & to\\
$\gamma_0$ & at & converge\\
 & convergence &\\
\hline
1.125   &   77.29\%   &   52628\\
 2.25   &   77.75\%   &   65811\\
  4.5   &   78.15\%   &   76209\\
  9.0   &   78.17\%   &   77235\\
\hline
\end{tabular}
\caption{\small Test accuracies at convergence and number of epochs to converge for different initial learning rates $\gamma_0$.
Low initial learning rates result in faster convergence to poorer local optimum.}
\label{tab:inception_lr}
\end{minipage}
\hspace{0.1cm}
\begin{minipage}[b]{0.64\textwidth}
\begin{subfigure}[t]{0.49\linewidth}
    \centering
    \includegraphics[width=1\linewidth]{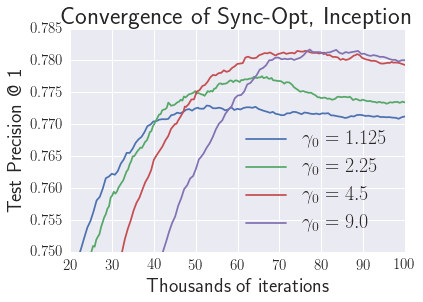}
    \caption{Convergence}
    \label{fig:inception_lr_converge}
\end{subfigure}
\begin{subfigure}[t]{0.49\linewidth}
    \centering
    \includegraphics[width=1\linewidth]{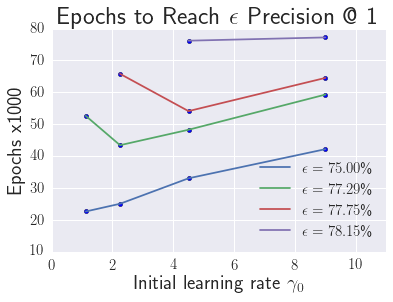}
    \caption{Epochs to $\epsilon$ test precision \@ 1.}
    \label{fig:inception_lr_itertoeps}
\end{subfigure}
\captionof{figure}{\small Convergence of Sync-Opt on Inception model using $N=100$ workers and $b=6$ backups, with varying initial learning rates $\gamma_0$.
To reach a lower $\epsilon$ test precision, small $\gamma_0$'s require fewer epochs than large $\gamma_0$'s.
However, small $\gamma_0$'s either fail to attain high $\epsilon$ precision, or take more epochs than higher $\gamma_0$'s.
}
\label{fig:inception_lr}
\end{minipage}
\end{table}

We ran Sync-Opt on Inception with $N=100$ and $b=6$, but varied the initial learning rate $\gamma_0$ between 1.125 and 9.0.
(Learning rates are exponentially decreased with iterations.)
Table \ref{tab:inception_lr} shows that smaller $\gamma_0$ converge faster, but to poorer test precisions.
Focusing on speed on an early phase of training could lead to misleading conclusions if we fail to account for eventual convergence.
For example, Figure \ref{fig:inception_lr_itertoeps} shows that using $\gamma_0=1.125$ reaches $\epsilon=75\%$ precision $1.5\times$ faster than $\gamma_0=4.5$, but is slower for $\epsilon=77.75\%$, and fails to reach higher precisions.

\subsection{Inception}
\label{sec:expt:inception}



\begin{figure*}[ht]
\centering
\begin{subfigure}[t]{0.49\linewidth}
    \centering
    \includegraphics[width=1\linewidth]{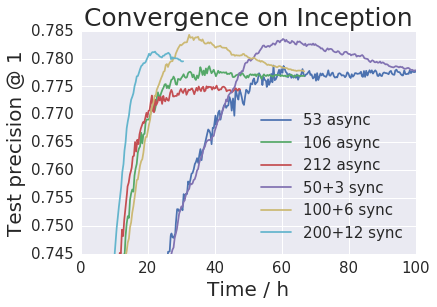}
    \caption{Convergence}
    \label{fig:inception_converge}
\end{subfigure}
\begin{subfigure}[t]{0.49\linewidth}
    \centering
    \includegraphics[width=1\linewidth]{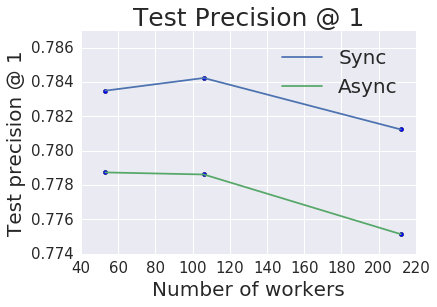}
    \caption{Test precision @ 1}
    \label{fig:inception_precision}
\end{subfigure}
\\
\begin{subfigure}[t]{0.32\linewidth}
    \centering
    \includegraphics[width=1\linewidth]{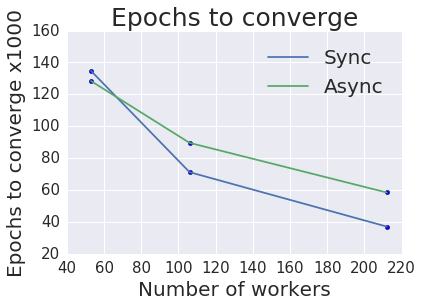}
    \caption{Epochs to converge}
    \label{fig:inception_epochs}
\end{subfigure}
\begin{subfigure}[t]{0.32\linewidth}
    \centering
    \includegraphics[width=1\linewidth]{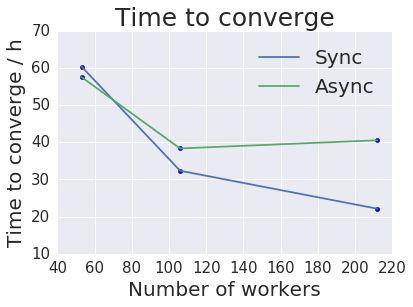}
    \caption{Time to converge}
    \label{fig:inception_time}
\end{subfigure}
\begin{subfigure}[t]{0.32\linewidth}
    \centering
    \includegraphics[width=1\linewidth]{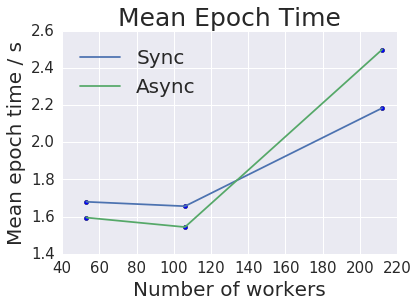}
    \caption{Mean epoch time}
    \label{fig:inception_epochtime}
\end{subfigure}
\caption{\small Convergence of Sync-Opt and Async-Opt on Inception model using varying number of machines.
Sync-Opt with backup workers converge faster, with fewer epochs, to higher test accuracies.}
\label{fig:inception}
\end{figure*}

We conducted experiments on the Inception model \citep{inception_v3} trained on ImageNet Challenge dataset \citep{imagenet},
where the task is to classify images out of 1000 categories. We used several configurations, varying $N+b$ from 53 to 212 workers.
Additional details of the training are provided in Appendix \ref{app:model:expt_inception}.
An {\em epoch} is a synchronous iteration for Sync-Opt, or a full pass of $N$ updates for Async-Opt, which represent similar amounts of computation.
Results of this experiment are presented in Figure \ref{fig:inception}.


Figure \ref{fig:inception_precision} shows that Sync-Opt outperforms Async-Opt in test precision:
Sync-Opt attains $\sim$0.5\% better test precision than Async-Opt for comparable $N+b$ workers.
Furthermore, Sync-Opt converges 6h and 18h faster than Async-Opt for 106 and 212 workers respectively, and is 3h slower when 53 workers are used, as seen in Figure \ref{fig:inception_time}.
This difference in speed is largely due to the fewer epochs (Figure \ref{fig:inception_epochs}) needed by Sync-Opt, but comparable or better epoch time (Figure \ref{fig:inception_epochtime}).

\subsection{PixelCNN Experiments}
\label{sec:expt:pixelcnn}
The second model we experimented on is PixelCNN \citep{oord2016conditional}, a conditional image generation deep neural network, which we train on the CIFAR-10 \citep{krizhevsky2009learning} dataset.
Configurations of $N+b=1,8,16$ workers were used;
for Sync-Opt, we always used $b=1$ backup worker.
Additional details are provided in Appendix \ref{app:model:expt_pixelcnn}.

\begin{figure*}[ht]
\centering
\begin{subfigure}[t!]{0.49\textwidth}
    \centering
    \includegraphics[width=1\textwidth]{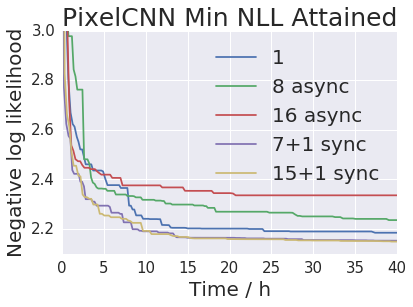}
    \caption{}
    \label{fig:pixelcnn_nll}
\end{subfigure}
\begin{subfigure}[t!]{0.49\textwidth}
    \centering
    \includegraphics[width=1\textwidth]{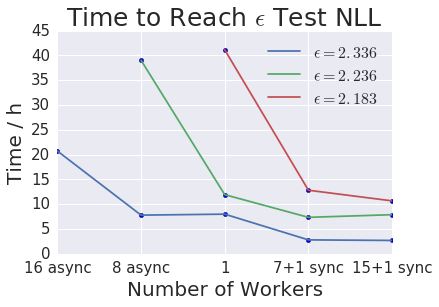}
    \caption{}
    \label{fig:pixelcnn_timenll}
\end{subfigure}
\caption{\small Convergence of synchronous and asynchronous training on PixelCNN model.
For clarity, we show the best NLL reached up to that point of time.
Sync-Opt achieves lower negative log likelihood in less time than Async-Opt.}
\label{fig:pixelcnn}
\end{figure*}

Convergence of the test negative log likelihood (NLL) on PixelCNN is shown in Figure \ref{fig:pixelcnn_nll}, where lower is better.
Observe that Sync-Opt obtains lower NLL than Async-Opt;
in fact, Async-Opt is even outperformed by serial RMSProp with $N=1$ worker, with degrading performance as $N$ increases from 8 to 16.
Figure \ref{fig:pixelcnn_timenll} further shows the time taken to reach $\epsilon$ test NLL.
Sync-Opt reduces the time to reach $\epsilon=2.183$ from 40h to $<13$h;
this NLL is not even achieved by Async-Opt.


\section{Related Work}
\label{sec:related}
Multicore and distributed optimization algorithms have received much attention in recent years.
Asynchronous algorithms include \cite{recht2011hogwild, duchi2013estimation, zhang2015deep, reddi2015variance, leblond2016asaga}.
Implementations of asynchronous optimization include \cite{xing2015petuum, li2014scaling, adam:2014}.
Attempts have also been made in \cite{zinkevich2010parallelized} and \cite{zhang2015splash} to algorithmically improve synchronous SGD.

An alternative solution, ``softsync'', was presented in \cite{zhang2015staleness},
which proposed batching gradients from multiple machines before performing an asynchronous SGD update,
thereby reducing the effective staleness of gradients.
Similar to our proposal, softsync avoids stragglers by not forcing updates to wait for the slowest worker.
However, softsync allows the use of stale gradients but we do not.
The two solutions provide different explorations of the trade-off between high accuracy (by minimizing staleness) and fast throughput (by avoiding stragglers).

\cite{watcharapichat2016ako} introduces a distributed deep learning system without parameter servers, by having workers interleave gradient computation and communication in a round-robin pattern.
Like Async-Opt, this approach suffers from staleness.
We also note that in principle, workers in Sync-Opt can double as parameter servers and execute the update operations and avoid the need to partition hardware resources between workers and servers.

\cite{das2016distributed} analyzes distributed stochastic optimization and optimizes the system by solving detailed system balance equations.
We believe this approach is complimentary to our work, and could potentially be applied to guide the choice of systems configurations for Sync-Opt.

\cite{keskar2016large} suggests that large batch sizes for synchronous stochastic optimization leads to poorer generalization.
Our effective batch size increases linearly with the number of workers $N$.
However, we did not observe this effect in our experiments; we believe we are not yet in the large batch size regime examined by \cite{keskar2016large}.

\section{Conclusion and Future Work}
\label{sec:conclude}
Distributed training strategies for deep learning architectures will become ever more important as the size of datasets increases.
In this work, we have shown how both synchronous and asynchronous distributed stochastic optimization suffer from their respective weaknesses of stragglers and staleness.
This has motivated our development of synchronous stochastic optimization with backup workers, which we show to be a viable and scalable strategy.

We are currently experimenting with different kinds
of datasets, including word-level language models where parts of the model (the embedding layers) are often
very sparse, which involves very different communication constraints.
We are also working on further improving the performance of synchronous
training like combining gradients from multiple workers sharing the same machine before
sending them to the parameter servers to reduce the communication overhead.
An alternative of using time-outs instead of backup workers is also being explored.

{\scriptsize
\bibliography{biblio}
\bibliographystyle{iclr2017_conference}
}

\newpage

\appendix

\section{Details of Models and Training}
\label{app:model}

\subsection{MNIST CNN, Section \ref{sec:asyncsgd:staleness}}
\label{app:model:mnistcnn}
The model used in our experiments is a 4-layer CNN that have 3x3 filters with
max-pooling and weight normalization in every layer. We trained the model with SGD
for 25 epochs and evaluated performance on the exponential moving average $\bar{\theta}$ using a decay rate of $\alpha = 0.9999$.
Initial learning rate was set to be 0.1 and linearly annealed to 0 in the last
10 epochs. We also used small image rotations and zooms as a data augmentation
scheme.

\subsection{Inception, Section \ref{sec:syncsgd:straggler}}
\label{app:model:straggler_inception}
For our straggler experiments, we trained the Inception \citep{inception_v3} model on the ImageNet Challenge dataset \citep{imagenet}.
10 parameter servers were used, and each worker was equipped with a k40 GPU.

The underlying optimizer was RMSProp with momentum, with decay of 0.9 and momentum of 0.9.
Mini-batch size $B=32$ was used.
Initial learning rates $\gamma_0$ were set at $0.045 N$, which we found to provide good test precisions for Inception.
Learning rates were also exponentially decreased with decay rate $\beta=0.94$ as $\gamma_0 \beta^{tN / (2T)}$, where $T=|\mathcal{X}| / B$ is the number of mini-batches in the dataset.

Test precisions were evaluated on the exponential moving average $\bar\theta$ using $\alpha = 0.9999$.

\subsection{Inception, Section \ref{sec:expt:inception}}
\label{app:model:expt_inception}
For experiments comparing Async-Opt and Sync-Opt on the Inception model in Section \ref{sec:expt:inception}, each worker is equipped with a k40 GPU.
For $N+b=53$ workers, 17 parameter servers were used; for $N+b=106$ workers, we used 27 parameter servers; and 37 parameter servers were used for $N+b=212$.

In the asynchronous training mode, gradient clipping is also needed for stabilization,
which requires each worker to collect the gradient across all layers of the deep model, compute the
global norm $||G||$ and then clip all gradient accordingly.
However, synchronization turns out to be very stable so gradient clipping is no longer needed, which
means that we can pipeline the update of parameters in different layers:
the gradient of top layers' parameters can be sent to parameter servers while concurrently computing gradients for the lower layers.

The underlying optimizer is RMSProp with momentum, with decay of 0.9 and momentum of 0.9.
Mini-batch size $B=32$ was used.
Initial learning rates $\gamma_0$ for Async-Opt were set to 0.045;
for Sync-Opt, we found as a rule-of-thumb that a learning rate of $0.045 N$ worked well for this model.
Learning rates were then exponentially decayed with decay rate $\beta=0.94$ as $\gamma_0 \beta^{t / (2T)}$ for Async-Opt, where $T=|\mathcal{X}| / B$ is the number of mini-batches in the dataset.
For Sync-Opt, we learning rates were also exponentially decreased at rate of $\gamma_0 \beta^{tN / (2T)}$, so that the learning rates after computing the same number of datapoints are comparable for Async-Opt and Sync-Opt.

Test precisions were evaluated on the exponential moving average $\bar\theta$ using $\alpha = 0.9999$.

\subsection{PixelCNN, Section \ref{sec:expt:pixelcnn}}
\label{app:model:expt_pixelcnn}
The PixelCNN \citep{oord2016conditional} model was trained on the CIFAR-10 \citep{krizhevsky2009learning} dataset.
Configurations of $N+b=1,8,16$ workers each with a k80 GPU, and 10 parameter servers were used.
For Sync-Opt, we always used $b=1$ backup worker.
The underlying optimizer is RMSProp with momentum, using decay of 0.95 and momentum of 0.9.
Initial learning rates $\gamma_0$ were set to $1e-4$ and slowly decreased to $3e-6$ after 200,000 iterations.
Mini-batch size $B=4$ was used.

\end{document}